%% file: main.tex
\documentclass[a4paper]{article}

\usepackage{INTERSPEECH2022}
\usepackage[backref=false]{hyperref}
\usepackage{multirow}
\hypersetup{plainpages = false,
              breaklinks=true,
              linktocpage,
              colorlinks   = true, 
              urlcolor     = blue, 
              linkcolor    = blue, 
              citecolor   = red, 
              pdfauthor={Niko Brummer},
              pdfstartpage=1, 
              pdfstartview=FitH,  
              pdfkeywords={}}

\title{Probabilistic Spherical Discriminant Analysis: An Alternative to PLDA for length-normalized embeddings}
\name{Niko Br\"ummer$^1$,
Albert Swart$^1$,
Ladislav Mo\v{s}ner$^{2}$,
Anna Silnova$^{2}$,
Old\v{r}ich Plchot$^{2}$,\\
Themos Stafylakis$^{3}$,
Luk\'a\v{s} Burget$^2$}
\address{
  $^1$Phonexia, South Africa\\
  $^2$Brno University of Technology, Speech@FIT and IT4I Center of Excellence, Brno, Czechia\\
	$^3$Omilia - Conversational Intelligence, Athens, Greece
	}
\email{niko.brummer@gmail.com, adswart@gmail.com}

\input{macros}

\def\textit#1{\emph{#1}}

\def\vmf{\mathcal{V}}

\begin{document}

\maketitle
\begin{abstract}
In speaker recognition, where speech segments are mapped to embeddings on the unit hypersphere, two scoring backends are commonly used, namely cosine scoring or PLDA. Both have advantages and disadvantages, depending on the context. Cosine scoring follows naturally from the spherical geometry, but for PLDA the blessing is mixed---length normalization Gaussianizes the between-speaker distribution, but violates the assumption of a speaker-independent within-speaker distribution. We propose PSDA, an analogue to PLDA that uses Von Mises-Fisher distributions on the hypersphere for both within and between-class distributions. We show how the self-conjugacy of this distribution gives closed-form likelihood-ratio scores, making it a drop-in replacement for PLDA at scoring time. All kinds of trials can be scored, including single-enroll and multi-enroll verification, as well as more complex likelihood-ratios that could be used in clustering and diarization. Learning is done via an EM-algorithm with closed-form updates. We explain the model and present some first experiments.
\end{abstract}

\section{Introduction}
Probabilistic \textit{linear} discriminant analysis (PLDA)~\cite{PLDA-IOFFE,PLDA-Prince}, is a popular backend for scoring speaker recognition embeddings in $\R^d$, following~\cite{Kenny_HTPLDA,NikoOdyssey10}. However,~\cite{Dani_length_norm} showed that length-normalizing the embeddings onto the unit sphere, $\S^{d-1}$ has a Gaussianizing effect that improves accuracy and this has been standard practice ever since. One disadvantage of the length-normalization is that within-speaker variability is squashed in the radial direction, making it \textit{speaker-dependent}, in violation of the PLDA assumption of a constant within-class distribution. Moreover, given a flexible, discriminatively trained embedding extractor, it is often found that cosine scoring (dot products between embeddings in $\S^{d-1}$) outperforms PLDA, especially when the test data is in domain, e.g.~\cite{SpeakIn_VoxSRC21,zeinali2019but}. We propose to enrich the field with a new backend that is intermediate between cosine scoring and PLDA: Probabilistic \textit{spherical} discriminant analysis (PSDA) uses Von Mises-Fisher (VMF) distributions on $\S^{d-1}$ in place of Gaussians. A Python implementation is available here.\footnote{\url{https://github.com/bsxfan/PSDA}}
   
To the best of our knowledge, the only related work is \cite{dubey2018robust}, where a VMF mixture was used for speaker clustering of length-normalized i-vectors. 

\section{Von Mises-Fisher distribution}
When embeddings in Euclidean space, $\R^d$ are length-normalized, they are projected onto the \emph{unit hypersphere}:\footnote{Do not confuse \textit{sphere} with \textit{ball}: $\S^{d-1}$ is the surface of the ball. Euclidean norm is denoted $\norm{\xvec}=\sqrt{\xvec'\xvec}$.}
\begin{align}
\S^{d-1} &= \{\xvec\in\R^d:\,\norm{\xvec}=1\}    
\end{align}
When $\xvec\in\S^{d-1}$, it is \emph{on} the sphere. If $\norm{\xvec}<1$, it is \emph{inside}. To construct the PSDA model, we replace the Gaussians in PLDA with Von Mises-Fisher (VMF) distributions. The density for $\xvec\in\S^{d-1}$ is~\cite{MardiaJupp}: 
\begin{align}
\vmf(\xvec\mid\muvec, \kappa) &= K_d C_\nu(\kappa)e^{\kappa\muvec'\xvec} &&\text{where} &
\nu&=\frac{d}{2}-1
\end{align}
The parameters are the \emph{mean direction}, $\muvec\in\S^{d-1}$ and the \emph{concentration}, $\kappa\ge0$. While $K_d$, depends only on the dimension and is of no further interest here,\footnote{Different authors (e.g.~\cite{MardiaJupp} vs~\cite{Banerjee05}) use different expressions for $K_d$, depending on the reference measure for the density. $K_d$ is analogous to the usually irrelevant $(2\pi)^{d/2}$ factor in the multivariate normal density.} the other normalization factor is all-important for our purposes:
\begin{align}
\label{eq:Cnu}
C_\nu(\kappa) &=\frac{\kappa^{\nu}}{I_\nu(\kappa)}  
= \Bigl(\sum_{i=0}^{\infty}\frac{\kappa^{2i}}{2^{2i+\nu}\, i!\,\Gamma(i+\nu+1)}\Bigr)^{-1}
\end{align}
where $I_\nu$ is the modified Bessel function of the first kind (Bessel-I) of order $\nu$. Note $I_\nu(\kappa)\ge0$ and $I_\nu(0)=0$ for $\nu>0$, but $I_0(0)=1$. The derivative can be expressed as~\cite{Banerjee05}:
\begin{align}
\label{eq:Ideriv}
\frac{\partial}{\partial\kappa}I_\nu(\kappa) &= \frac{\nu}{\kappa}I_\nu(\kappa)+I_{\nu+1}(\kappa)
\end{align}
which shows it is monotonic rising. As a function of $\kappa$, $C_\nu(\kappa)$ is positive and strictly monotonic decreasing,\footnote{On a log-log plot it is relatively flat for $0\le\kappa<<\sqrt{\nu+1}$ and then plunges dramatically for large $\kappa$.} and $\lim_{\kappa\to0}C_\nu(\kappa)=2^\nu\Gamma(\nu+1)$. The concentration parameter, $\kappa$ is roughly analogous to precision in the normal distribution. For smaller $\kappa$, the distribution is more widely spread, until at $\kappa=0$ it gives the uniform hypersphere distribution. For larger $\kappa$, the distribution concentrates more tightly around $\muvec$. It should be noted that $\muvec\in\S^{d-1}$ is \textit{not the expected value}, which instead is at~\cite{MardiaJupp}:  
\begin{align}
\label{eq:ev}
\expv{\xvec}{} &= \rho(\kappa)\muvec, &&\text{where} &
0\le\rho(\kappa) &= \frac{I_{\nu+1}(\kappa)}{I_\nu(\kappa)} < 1
\end{align}
where $\expv{\xvec}{}$ is inside the sphere, not on it.
The norm, $\norm{\expv{\xvec}{}}=\rho(\kappa)$ is strictly increasing w.r.t.\ $\kappa$, where $\lim_{\kappa\to0}\rho(\kappa)=0$ and $\lim_{\kappa\to\infty}\rho(\kappa)=1$. 
The empirical mean, say $\bar\xvec$, of a cluster of points on the hypersphere has the same behaviour: $\bar\xvec$ is inside the sphere and moves closer to it ($\norm{\bar\xvec}$ increases towards $1$), as the cluster becomes more concentrated.  

\subsection{Maximum likelihood parameter estimates}
Given data set in $\S^{d-1}$, say $\Xmat=\{\xvec_i\}_{i=1}^n$, assumed to have been sampled \emph{iid} from $\vmf(\muvec,\kappa)$, the maximum-likelihood (ML) estimate of the parameters is obtained by maximizing the log-likelihood:
\begin{align}
\label{eq:MLobj}
\log\prod_{i=1}^n \vmf(\xvec_i\mid\muvec,\kappa) = n\log C_\nu(\kappa) + n\kappa\muvec'\bar\xvec +\const
\end{align}
where $\bar\xvec=\frac1n\sum_i\xvec_i$. The maximum w.r.t.\ $\muvec\in\S^{d-1}$ is at:
\begin{align}
\muvec_{ML}&=\frac{\bar\xvec}{\norm{\bar\xvec}}
\end{align}
Inserting this back into~\eqref{eq:MLobj}, we need to maximize:  
\begin{align}
\label{eq:MLobj2}
n\log C_\nu(\kappa) + n\kappa\norm{\bar\xvec}
\end{align}
to find $\kappa$. Setting the derivative to zero, using~\eqref{eq:Ideriv}, gives~\cite{MardiaJupp}:
\begin{align}
\kappa_{ML}&=\rho^{-1}(\norm{\bar\xvec})
\end{align}
We used a numerical (derivative-free) rootfinding algorithm\footnote{\texttt{scipy.optimize.toms748}} to invert $\rho$. At the ML estimate $\expv{\xvec}{}=\bar\xvec$. The ML parameters depend \textit{solely} on the sufficient statistic $\bar\xvec$. 
When $\bar\xvec=\nulvec$, $\muvec$ is irrelevant and the likelihood is maximized at $\kappa=0$, which gives the uniform distribution.

\section{The PSDA model}
PSDA is constructed much like PLDA~\cite{NikoOdyssey10}. For every speaker we posit a hidden speaker identity variable, $\zvec\in\S^{d-1}$, having a VMF prior, $\vmf(\zvec\mid\muvec,b)$, where $\muvec\in\S^{d-1}$ is the \emph{speaker mean direction} and $b\ge0$ is the \emph{between-speaker concentration}. Embeddings with low speaker concentration (as spread out as possible), ideally $b=0$, is required for good accuracy. The simplest variant of PSDA has a uniform between-speaker distribution ($b=0$ and $\muvec$ irrelevant). If however, the speaker distribution is believed to be non-uniform, $b$ and $\muvec$ can be learnt from labelled data, as we are accustomed to do with PLDA. 

The observed embeddings in $\S^{d-1}$, are supposed to have been generated from speaker-dependent VMF distributions: embeddings from different speakers are independent and those from the same speaker are conditionally independent, given $\zvec$. If $\Xmat=\{\xvec_i\}_{i=1}^n$ are embeddings of a common speaker, then:
\begin{align}
P(\Xmat\mid\zvec) &= \prod_i \vmf(\xvec_i\mid\zvec,w) \propto \exp\bigl[w\zvec'\bar\xvec\bigr]
\end{align}
where $\bar\xvec=\frac1n\sum_i\xvec_i$ and where $w>0$ is the \emph{within-speaker concentration}. Note the conjugacy: the product of VMF distributions for the observed data doubles as a likelihood function for $\zvec$, which is also in VMF form. 

In summary, the learnable PSDA model parameters are $w,b\in\R$ and $\muvec\in\S^{d-1}$. Next we show how to do inference and learning. We start with inference for $\zvec$, followed by inference for the speaker hypothesis (scoring). Finally, maximum-likelihood learning can be done with an EM-algorithm.

\subsection{The hidden variable posterior}
Given one or more observations, $\Xmat=\{\xvec\}_{i=1}^n$, assumed to be of the same speaker, the identity variable posterior is still VMF:
\begin{align}
\label{eq:zpost}
\begin{split}
P(\zvec\mid\Xmat) &\propto \vmf(\zvec\mid\muvec,b) \prod_i\vmf(\xvec_i\mid\zvec,w) \\
&\propto \exp\Bigl[\Bigl(b\muvec+w\sum_i\xvec_i\Bigr)'\zvec\Bigr] \\
&\propto\vmf\Bigl(\zvec\Bigm|\frac{\tilde\zvec}{\norm{\tilde\zvec}},\norm{\tilde\zvec}\Bigr)
\end{split}
\end{align}
where $\tilde\zvec = b\muvec+w\sum_i\xvec_i$. The concentrations, $b$ and $w$ behave in much the same way as the precisions in Gaussian PLDA. But, in Gaussian PLDA~\cite{NikoOdyssey10}, the posterior precision is dependent only on the number of observations, while here, the posterior concentration, $\norm{\tilde\zvec}$ is data-dependent. If the data all lie in the same quadrant then the more data we have, the more the concentration will grow. But if the data are spread with angles wider than 90 degrees, they can (partially) cancel and the posterior concentration can become arbitrarily small. (E.g.~ if $b=0$ and there are two antipodal observations, then $\tilde\zvec=\nulvec$.) This stands in contrast to the heavy-tailed PLDA model of~\cite{HTPLDA}, where larger norms are associated with larger within-speaker variation.

\subsection{Scoring}
Given a trained PSDA model, with parameters $(w,b,\muvec)$, we derive a general recipe for computing likelihood-ratio scores. As with PLDA~\cite{NikoOdyssey10}, PSDA provides closed-form scores for a variety of verification and clustering trials. Let $\Emat=\{\evec_1,\ldots,\evec_m\}$ denote a collection of $m\ge1$ enrollment observations hypothesized to be from a common speaker. $\Tmat=\{\tvec_1,\ldots,\tvec_n\}$ denotes a collection of $n\ge1$ test observations from a common (but possibly different) speaker. We want to compute a \textit{likelihood-ratio} (LR) for hypothesis $H_1$, that all observations come from one common speaker, against hypothesis $H_2$, that they come from two different speakers:
\begin{align}
\label{eq:LR}
\frac{P(\Emat,\Tmat\mid H_1)}{P(\Emat,\Tmat\mid H_2)} &= \frac{P(\Emat,\Tmat\mid H_1)}{P(\Emat\mid H_1)P(\Tmat\mid H_1)} 
\end{align}
All RHS factors are marginals, where $\zvec$ has been integrated out. We can use the conjugacy and the availability of the VMF normalizer~\eqref{eq:Cnu} to solve these integrals in closed form. Following the derivation in~\cite{NikoOdyssey10}, the LR can be rewritten as:
\begin{align}
\frac{P(\Emat,\Tmat\mid H_1)}{P(\Emat,\Tmat\mid H_2)} 
&= \frac{P(\zvec_0\mid\Emat)P(\zvec_0\mid\Tmat)}
{P(\zvec_0\mid\Emat,\Tmat)P(\zvec_0)}
\end{align}
Since the LHS is independent of $\zvec_0\in\R^{d-1}$, so is the RHS: all factors of the form $e^{\zvec_0'\cdots}$ cancel, leaving only the VMF normalization constants to yield our general \emph{closed-form scoring function}:
\begin{align}
\label{eq:score}
\frac{P(\Emat,\Tmat\mid H_1)}{P(\Emat,\Tmat\mid H_2)} 
&= \frac{C_\nu(\norm{b\muvec+w\tilde\evec})C_\nu(\norm{b\muvec+w\tilde\tvec})}
{C_\nu(\norm{b\muvec+w\tilde\evec+w\tilde\tvec})C_\nu(b)}
\end{align}
where $\tilde\evec=\sum_{i=1}^m \evec_i$, and $\tilde\tvec=\sum_{i=1}^n \tvec_i$.

\subsubsection{Relationship with cosine scoring}
In the special case when we set the model parameter $b=0$ and when the enroll and test sets are singletons, $m=n=1$, there is a close relationship between the PSDA score~\eqref{eq:score} and the ubiquitous cosine score. When $b=0$, the score simplifies to:
\begin{align}
\frac{P(\Emat,\Tmat\mid H_1)}{P(\Emat,\Tmat\mid H_2)} 
&= \frac{C_\nu(w)^2}
{C_\nu(w\norm{\tilde\evec+\tilde\tvec})\,\lim_{b\to0}C_\nu(b)}
\end{align}
When $\tilde\evec=\evec_1\in\S^{d-1}$ and $\tilde\tvec=\tvec_1\in\S^{d-1}$, the \textit{cosine score} is the dot product $\left<\tilde\evec,\tilde\tvec\right>$, which can be rewritten as: 
\begin{align}
\left<\tilde\evec,\tilde\tvec\right>  
&= \frac{\norm{\tilde\evec+\tilde\tvec}^2-2}{2}
\end{align}
Since $C_\nu$ is monotonic decreasing, there is a monotonic rising functional relationship between the PSDA score and the cosine score. This means the EER and minDCF and in general the whole DET-curve will be identical for cosine scoring and PSDA. However, for $b>0$ and also all other kinds of trials, this scoring formula~\eqref{eq:score} has a more complex and possibly more useful behaviour.

The proposed model can be seen as intermediate between cosine scoring and PLDA. Cosine scoring has no learnable parameters, while PLDA has a substantial set of parameters, on the order of $d^2$. The PSDA model has parameters, but only about $d$ of them. When $b$ is small the EER for PSDA can get arbitrarily close to cosine scoring, but some extra flexibility is obtained when $w,b,\muvec$ are trained. For multiple enrollments, this model also gives a more interesting (arguably more principled) computation compared to simple averaging of the enrollment embeddings.  
 
\subsubsection{Scoring implementation}
To implement~\eqref{eq:score} we note some details. Bessel-I functions are tricky. 
$I_\nu(\kappa)$ is available in \texttt{scipy.special}, but when $\nu$ is large (as here), both overflow and underflow occur. Overflow can be managed by using \texttt{scipy.special.ive}, which implements $I_\nu(\kappa)e^{-\kappa}$, which we used for:
\begin{align}
\log I_\nu(\kappa) &= \log\bigl(I_\nu(\kappa)e^{-\kappa}\bigr) + \kappa
\end{align}
This still underflows for small $\kappa$. 
Whenever $\kappa<\sqrt{\nu+1}$, we used the first few terms (say 5) of the series expansion:
\begin{align}
\log I_\nu(\kappa) &= \log \sum_{i=0}^{\infty}\frac{(\kappa/2)^{2i+\nu}}{\Gamma(i+1)\Gamma(i+\nu+1)} 
\end{align}
using \texttt{logsumexp}, $\log\kappa$ and \texttt{gammaln}. 

Since $\norm{b\muvec+w\tilde\evec+w\tilde\tvec}$ is required in the denominator of~\eqref{eq:score}, for \textit{every} trial, a fast implementation is desirable. The numerator norms are of lesser concern. We can rewrite the norm as:
\begin{align}
\begin{split}
&\norm{b\muvec+w\tilde\evec+w\tilde\tvec}^2 \\
&= \norm{b\muvec+w\tilde\evec}^2 + \norm{w\tilde\tvec}^2 + 2\left<b\muvec+w\tilde\evec,w\tilde\tvec\right>
\end{split}
\end{align}
where the dot product for an $m$-by-$n$ block of scores can be implemented with a fast matrix multiplication.

\subsection{Learning}
Maximum-likelihood-learning can be done via an EM algorithm with closed-form updates, given labelled observations for a number of speakers. For each speaker, $i$, let there be $n_i$ observations with mean, $\bar\xvec_i$. The required statistics are just zero and first-order stats. In Gaussian PLDA, we need second-order statistics of the data too, but here, the speaker cluster spread is effectively contained in $\norm{\bar\xvec_i}$. The total number of observations is $N=\sum_i n_i$ and the number of training speakers is $S$. The E-step is the computation of the \emph{EM auxiliary}:
\begin{align}
\begin{split}
& Q(w,b,\muvec) = \const +\\
&\;\;\sum_i \expv{\log P(\Xmat_i\mid\zvec,w) + \log P(\zvec\mid\muvec,b)}{P(\zvec\mid \Xmat_i)} \\
&= \;\;\sum_i \expv{n_i\log C_\nu(w)+n_i w\bar\xvec_i'\zvec}{P(\zvec\mid \Xmat_i)} +\\
&\;\;\;\;\;\;\;\;\;\;\;\;\expv{\log C_\nu(b)+b\muvec'\zvec}{P(\zvec\mid \Xmat_i)} \\
&=  \;\;N\log C_\nu(w)+Nw\frac1N\sum_i n_i\bar\xvec_i'\expv{\zvec}{i} +\\
&\;\;\;\;\;\;\;S\log C_\nu(b)+Sb\muvec'\frac1S\sum_i\expv{\zvec}{i} 
\end{split}
\end{align}
The expectations are taken w.r.t.\ the posteriors~\eqref{eq:zpost}, where $\expv{\zvec}{i}$ is the posterior expectation for speaker $i$, as given by~\eqref{eq:ev}. The M-step maximizes $Q$ w.r.t.\ the parameters. For $b,\muvec$ this can be done by identifying the last line above with~\eqref{eq:MLobj}. For $w$, we identify the second last line with~\eqref{eq:MLobj2}. This gives the updates:
\begin{align}
\muvec &\gets \frac{\bar\zvec}{\norm{\bar\zvec}}, & 
b &\gets \rho^{-1}(\norm{\bar\zvec}), &
w &\gets \rho^{-1}(\norm{\bar\rvec})
\end{align} 
where: 
\begin{align}
\bar\zvec&=\frac1S\sum_i\expv{\zvec}{i}, &&\text{and} &
\norm{\bar\rvec}&=\frac1N\sum_i n_i\bar\xvec_i'\expv{\zvec}{i}
\end{align}
Since $\rho^{-1}$ is monotonic rising, we see:
\begin{itemize}
\item When $\bar\zvec$ is closer to the origin, the estimated between-speaker concentration, $b$ is smaller (good for accuracy).
\item The better the observations align with the hidden variables, the higher the estimated within-speaker concentration, $w$ (also good).
\end{itemize}


\def\tb#1{\textbf{#1}}
\begin{table*}[htbp!]
    \centering
    \caption{Comparison of the speaker verification performance for cosine scoring, PLDA, and PSDA. The performance metrics are EER (\%) and minDCF for $p_{\mathrm{tar}}=0.05$}
    \begin{tabular}{l l c  c c c c c }
        \toprule
        & \multirow{2}{*}{\tb{Back-End}} & \multirow{2}{*}{\tb{Dim.reduction}} & 
        \multicolumn{2}{c}{\tb{VoxCeleb1-O}} &  \multicolumn{2}{c}{\tb{VoxCeleb1-H}} \\
         & & & \tb{EER} & \tb{MinDCF} & \tb{EER} & \tb{MinDCF}\\
        \midrule
        &cos&-&1.10&0.071&2.13&0.126\\
        &cos&PCA 100&1.12&0.073&2.21&0.130\\
        &cos&LDA 100&2.05&0.157&5.28&0.340\\
        &PLDA&-&4.47&0.254&6.80&0.325\\
        &PLDA&PCA 100&1.34&0.094&2.69&0.152\\
        &PLDA&LDA 100&2.16&0.157&4.61&0.273\\
         &PSDA&-&1.10&0.071&2.13&0.126\\
         &PSDA&PCA 100& 1.12&0.073&2.21&0.130\\
          &PSDA&LDA 100&2.04&0.155&5.20&0.333\\
        \bottomrule
    \end{tabular}
    \label{tab:results}
\end{table*}

\section{The VMF classification head}
The VMF-based PSDA model provides some interesting insights into embedding extractor training. The functional form of the standard discriminatively trained multiclass linear classifier, with inputs in $R^d$, is obtained by letting the logits (softmax inputs) be the class log-likelihoods of a Gaussian model with a common within-class covariance. The logit for class $i$ is: 
\begin{align}
\log \ND(\xvec\mid\muvec_i,\Pmat^{-1})=\muvec_i'\Pmat\xvec-\frac12\muvec_i'\Pmat\muvec_i+\const
\end{align}
where $\xvec,\muvec_i\in\R^d$. Note the \textit{class-dependent offsets}. 
If instead, we restrict $\xvec,\muvec_i\in\S^{d-1}$, we can derive a similar classifier using a VMF model, with common within-class concentration, so that the logits become:
\begin{align}
\label{eq:vmf_logit}
\log \vmf(\xvec\mid\muvec_i,\kappa)=\kappa\muvec_i'\xvec+\const    
\end{align}
now \textit{without} the offsets.\footnote{If there is a known class imbalance, fixed, non-trainable offsets $\log p_i$, where $p_i$ is class proportion, can be added to the logits.} Although logits of the form~\eqref{eq:vmf_logit} are now almost ubiquitous in machine learning,\footnote{$\kappa^{-1}$ is usually termed \textit{temperature}} we are not aware that the connection with VMF likelihood has been noted before. In speaker recognition, when embeddings extractors are trained classifier-style, with one class per training speaker, \eqref{eq:vmf_logit} is also used, although the problem may be made artificially harder by modifying the target logits with a margin, as in AM-softmax~\cite{wang2018cosface} and AAM-softmax~\cite{deng2019arcface}.

In a variety of embedding extractors trained with AM-softmax and AAM-softmax, we found~\cite{BUTOdyssey22} that the length-normalized embeddings tend to collapse almost to a subspace, with very little variability in at least half of the dimensions~\cite{LeCunCollapse22}. Curiously, if we train 512-dimensional embeddings, they collapse to less than 256 dimensions. If we train 256-dimensional embeddings, they collapse to less than 128 dimensions. The collapsed data is not modelled by PSDA, and we had mixed results with using PCA and LDA dimensionality reduction to better fit the data to the model. For future work, we are interested in modifying the embedding extractor training criterion to combat such collapse and instead encourage a uniform hypersphere between-speaker distribution. The ideas in~\cite{Isola, LeCunCollapse22} may be helpful.

\section{Experiments}
In our experiments, we compare PSDA with two baseline back-ends: PLDA and cosine scoring. 
The experimental setup is as follows:
256-dimensional embeddings were extracted with ResNet34 that was trained on the development part of VoxCeleb2 dataset~\cite{Nagrani19} to optimize AAM loss.
We used the same training dataset to train the back-end. The only difference was that for training the back-end we concatenate all segments belonging to the same session before extracting the embedding, while for training the extractor the original VoxCeleb segments were used. Also, we do not use any augmentations when training the back-end. The performance is tested on two conditions: original test set of VoxCeleb1 (VoxCeleb1-O) and on the ``hard'' trial list from VoxCeleb1 (VoxCeleb1-H). We report the performance in terms of equal Error Rate (EER) and minimum Detection Cost Function for the prior of target trial set to 0.05.  

As commented before, we have noticed that roughly one half of the dimensions of the embeddings have almost no variability. Thus, we want to perform dimensionality reduction before training the back-end model. In the experiments, we compare PCA, LDA, and using no dimensionality reduction for each of the three back-ends.
The results are presented in Table~\ref{tab:results}.
There are a few observations that we want to point out.
First, in our experiments, cosine scoring provides better performance than PLDA model for any embedding pre-processing. This finding agrees with previous works (see e.g.~\cite{zeinali2019but}). Notice that for PLDA trained on the embeddings without dimensionality reduction we had to set the size of the speaker and channel subspaces to 100. Otherwise, the training crashed. In other cases, we use two-covariance PLDA i.e. speaker and channel hidden variables have the same dimensionality as the observed data. 
Second, we observe that for both cosine scoring and PSDA, the raw embeddings without dimensionality reduction provide better performance than using low-dimensional embeddings. For PLDA we see the opposite trend: for good performance PLDA has to be trained on the embeddings after dimensionality reduction. 
Finally, we notice that the results of cosine scoring and PSDA are very similar for the embeddings with and without dimensionality reduction with the best performance achieved when they are used with raw embeddings. 

\section{Conclusion}
In speaker recognition state of the art (as in many other machine learning problems), length-normalized embeddings empirically perform better. Cosine scoring follows naturally, but is merely geometrically motivated (within-class distances should be small, between-class large). The PLDA model provides a beautiful, rich probabilistic scoring recipe, but makes use of Gaussians to model densities in $\R^d$. However, distributions restricted to $\S^{d-1}$ do not even possess densities in $\R^d$. Gaussians can only approximately fit length-normalized data. The problem of speaker-dependent within-class distributions has been mentioned above. We have shown that by using VMF distributions instead, PSDA can model distributions directly in $\S^{d-1}$, while still enjoying the scoring and training benefits of PLDA. We have shown theoretically and empirically, that (up to calibration) PSDA can be equivalent to cosine scoring. We hope this new tool can provide new theoretical insights and practical tools for both embedding extraction and scoring algorithms. In future we aim to explore calibration properties of PSDA, mixtures of PSDA, and PSDA scoring for clustering and diarization.

\bibliographystyle{IEEEtran}
\bibliography{mybib}

\end{document}

%% file: macros.tex
\def\zvec{\mathbf{z}}

\def\evec{\mathbf{e}}
\def\tvec{\mathbf{t}}

\def\ND{\mathcal{N}}

\def\expv#1#2{\left\langle#1\right\rangle_{#2}}

\def\R{\mathbb{R}}

\def\Emat{\mathbf{E}}
\def\Imat{\mathbf{I}}

\def\Xmat{\mathbf{X}}

\def\Tmat{\mathbf{T}}

\def\Pmat{\mathbf{P}}

\def\xvec{\mathbf{x}}

\def\rvec{\mathbf{r}}

\def\muvec{\boldsymbol{\mu}}

\def\nulvec{\boldsymbol{0}}

\def\const{\text{const}}

\def\S{\mathbb{S}}

\newcommand{\inv}[1]{%
\ifx{#1}{\Imat}           
  #1
\else
  {#1}^{-1}
\fi
}

\def\norm#1{\left\|#1\right\|}

\usepackage{tikz}
\usetikzlibrary{bayesnet}

\usetikzlibrary{arrows,shapes,backgrounds,positioning,fit}

\tikzstyle{cbox} = [rectangle,draw=blue!100,thick,align=center,rounded corners = 3pt]
\tikzstyle{lbox} = [rectangle,draw=blue!100,thick,align=left,rounded corners = 3pt]
\tikzstyle{ccircle} = [circle,draw=blue!100,thick,align=center,inner sep = 0]
\tikzstyle{ctext} = [rectangle,align=center,inner sep = 4pt]
\tikzstyle{ltext} = [rectangle,align=left,inner sep = 4pt]
\tikzstyle{solder} = [circle,draw,fill,inner sep = 0, minimum size = 3pt]

\def\textit#1{\emph{#1}}